\documentclass[3p,12pt]{elsarticle}

\usepackage{amssymb}
\usepackage[utf8]{inputenc}
\usepackage{lscape}
\usepackage{longtable}
\usepackage{makecell}
\usepackage{amsmath}
\usepackage{multirow}

\journal{arXiv}

\begin{document}

\begin{frontmatter}

\title{Soccer line mark segmentation and classification with stochastic watershed transform}

\author[inst1]{Daniel Berjón}

\affiliation[inst1]{organization={Information Processing and Telecommunications Center --- Universidad Polit\unexpanded{\'e}cnica de Madrid},
            addressline={Av. Complutense, 30}, 
            city={Madrid},
            postcode={28029}, 
            country={Spain}}

\author[inst1]{Carlos Cuevas}
\author[inst1]{Narciso García}

\begin{abstract}
Augmented reality applications are beginning to change the way sports
are broadcast, providing richer experiences and valuable insights
to fans. The first step of augmented reality systems is camera calibration,
possibly based on detecting the line markings of the playing field.
Most existing proposals for line detection rely on edge detection
and Hough transform, but radial distortion and extraneous edges cause
inaccurate or spurious detections of line markings. We propose a novel
strategy to automatically and accurately segment and classify line
markings. First, line points are segmented thanks to a stochastic
watershed transform that is robust to radial distortions, since it
makes no assumptions about line straightness, and is unaffected by
the presence of players or the ball. The line points are then linked
to primitive structures (straight lines and ellipses) thanks to a
very efficient procedure that makes no assumptions about the number
of primitives that appear in each image. The strategy has been tested
on a new and public database composed by 60 annotated images from
matches in five stadiums. The results obtained have proven that the
proposed strategy is more robust and accurate than existing approaches,
achieving successful line mark detection even in challenging conditions.
\end{abstract}


\end{frontmatter}

\global\long\def\imagen{I}%
\global\long\def\maskgrayscale{\imagen_{\mathrm{GS}}}%
\global\long\def\maskini{\imagen_{\mathrm{i}}}%

\global\long\def\randomWatershedImage{\imagen_{\mathrm{RWS}}}%
\global\long\def\openImage{\imagen_{\mathrm{open}}}%

\global\long\def\mascara{M}%

\global\long\def\mascaracampo{\mascara_{\mathrm{PF}}}%

\global\long\def\cosaboundary#1{#1_{\mathrm{B}}}%

\global\long\def\cosatophat#1{#1_{\mathrm{TH}}}%

\global\long\def\mascaralineas{\cosaboundary{\mascara}}%
\global\long\def\opening{e_{\mathrm{sd}}}%
\global\long\def\chromadistortion{\mathrm{cd}}%

\global\long\def\seedset{S}%
\global\long\def\numseeds{N}%
\global\long\def\numexperiments{M}%
\global\long\def\seedindex{j}%
\global\long\def\seedindexrow{j}%
\global\long\def\seedindexcol{k}%
\global\long\def\expindex{i}%

\global\long\def\myfunc#1#2{#1{\left(#2\right)}}%

\global\long\def\normald#1#2{\myfunc{\mathcal{N}}{#1,#2}}%

\global\long\def\uniformrv#1#2{\myfunc{\mathcal{U}}{#1,#2}}%
\global\long\def\seededWS#1#2{\myfunc{\mathrm{SeededWS}}{#1,#2}}%

\global\long\def\modulo#1{\myfunc{\mathrm{mod}}{#1}}%
\global\long\def\hessian{\mathbf{H}}%
\global\long\def\hessianof#1{\myfunc{\hessian}{#1}}%

\global\long\def\individualseed{\mathbf{s}}%
\global\long\def\rowcoord{r}%
\global\long\def\colcoord{c}%

\global\long\def\thresholdRWS{T_{\mathrm{RWS}}}%

\global\long\def\numseedsrows{\numseeds_{\mathrm{\rowcoord}}}%
\global\long\def\numseedscols{\numseeds_{\mathrm{\colcoord}}}%

\global\long\def\alto{H}%

\global\long\def\ancho{W}%

\global\long\def\recall{\mathrm{rec}}%
\global\long\def\precision{\mathrm{pre}}%

\global\long\def\fscore{F}%
\global\long\def\truepos{\mathrm{tp}}%

\global\long\def\falsepos{\mathrm{fp}}%
\global\long\def\falseneg{\mathrm{fn}}%

\section{Introduction\label{sec:Introduction}}

Soccer is the most popular sport in the world, not
only in terms of television audience (almost 4 billion followers in
200 countries) or players (more than 260 million)~\cite{Felix2020},
but also regarding research~\cite{kirkendall2020evolution}. Much
of the research work proposed in recent years has focused on responding
to the demand for applications capable of enriching the content of
live broadcasts with augmented reality~\cite{goebert2020new} and
applications aimed at analyzing and understanding the game~\cite{andrienko2019constructing,kapela2015real}.

To tackle these high-level tasks, it is required to register the images
in a model of the playing field~\cite{cuevas2020automatic} and/or
calibrate the cameras~\cite{chen2018two,citraro2020real}. The strategies with these
aims are typically based on the detection of key-points determined
from the intersections between the line marks on the grass~\cite{yao2017fast}.
These line marks can be of two types, straight lines and circles (seen
as ellipses in the images). Therefore, the location of the line marks
is a key stage in all these high-level tasks. However, existing methods
exhibit shortcomings in difficult lighting conditions, require manual tuning
or rely on assumptions about the straightness or distribution of line
markings.

In this paper we propose a novel method to detect the playing field line
points and determine which of them belong to straight lines and which
to ellipses. The quality of the results obtained has been assessed
in a database composed of numerous annotated images taken from a wide 
variety of points of view, in different and challenging lighting conditions, 
and its usefulness
has been demonstrated after being compared with other
state-of-the-art methods. Although we focus on soccer videos, the
proposed strategy can be adapted to other types of \textquotedblleft pitch
sports\textquotedblright .

\subsection{Contribution}

The main contributions of this work are:
\begin{itemize}
\item Line mark segmentation using a stochastic watershed transformation.
Many proposals for line marking detection start from a proto-edge-detection
stage that yields many unwanted or duplicate edges. Unlike existing methods, we
use the watershed transformation because it lends itself naturally
to flood out irrelevant edges such as those due to the players or
the ball.
\item Seed placing strategy. Our proposal builds upon the stochastic watershed
algorithm~\cite{angulo2007stochasticWatershed}, but our goal is
different (line marking detection instead of region segmentation);
hence, our method does not require the user to manually set a number of 
relevant regions to be sought and ensures quick convergence.
\item Viewpoint-independent classification of line marks into the two 
kinds of primitive structures
that appear in soccer fields: straight lines and ellipses. Previous
works apply independent algorithms to detect straight lines and ellipses,
while we perform a joint analysis that provides higher quality results.
Additionally, unlike those works, we not only detect the center circle
but also the penalty arcs.
\end{itemize}

\subsection{Organization}

The paper is organized as follows: in section~\ref{sec:Related-work}
we review existing algorithms for line marking detection. 
Sections~\ref{sec:Watershed-segmentation} and~\ref{sec:Line-classification} describe 
our proposal for line mark segmentation and classification, respectively. 
Experiment results are reported in section~\ref{sec:Results}
and, finally, section~\ref{sec:Conclusions} presents the conclusions
of the paper.

\section{Related work\label{sec:Related-work}}

To measure distances and speeds of players and/or ball throughout
the matches, the images need to be registered to a model of the playing
field. This is typically accomplished from sets of key-points that
result from intersections between line marks. 

To bring out the white line marks from the rest of the elements in
the playing field, some authors use edge detectors: the Sobel detector
in~\cite{rao2015novel,zhang2015research}, the Canny detector in~\cite{direkoglu2018player,doria2021soccer},
or the Laplacian of Gaussian (LoG) detector in~\cite{szenberg2001automatic,bu2011automatic}.
To obtain the points centered in the line marks, other authors use
the Top-Hat transform~\cite{cuevas2020automatic,sun2009field,yang2017robust}.
There are also works that apply combinations of morphological operations~\cite{aleman2014camera}.
All these methods, although quick and simple, have the important drawback
of the correct selection of a threshold. If the threshold value is
too high, they are not able to detect the lines in the lower contrast
areas of the image (heavily shadowed or brightly lit areas). On the
other hand, if the threshold is too low, numerous false detections
occur due to the grass texture and the presence of players on the
pitch.

\begin{figure}[tbh]
\centering{}\includegraphics[width=1\columnwidth]{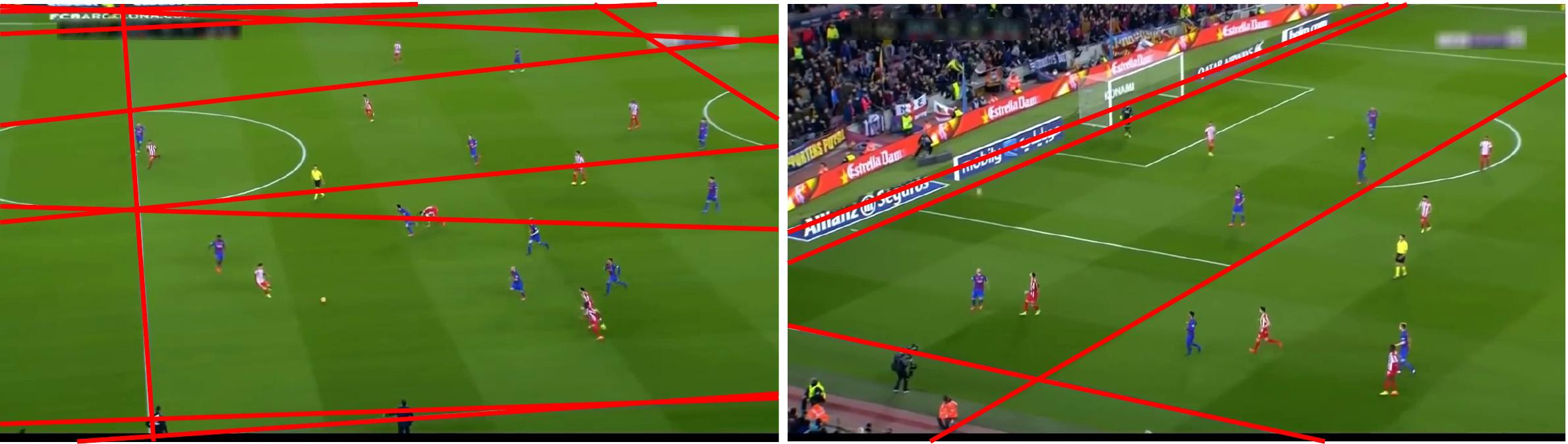}\caption{\label{fig:Related}Examples of typical line mark detections using the Hough transform superimposed on
the original images.}
\end{figure}

Once the white lines mask is obtained, most strategies apply
the Hough transform to detect the straight line marks~\cite{yao2017fast}.
The Hough transform is computationally efficient and provides successful
results in simple images. However, it is very sensitive to the presence
of the aforementioned false detections and therefore results in numerous
false lines. An additional drawback of Hough-based methods is the
need to set the number of lines to consider. Typically, in images
captured by the Master Camera\footnote{The master camera is the one used 
most of the time in soccer broadcasting and also the only one considered 
in most approaches in the literature.
It is placed approximately on the extension of the halfway line. It
performs pan, tilt, and zoom movements, but not rotations around its
longitudinal axis (i.e., no roll).}, the maximum number of straight lines 
that can be seen is 10 (straight
lines on each of the halves of the playing field). However, in many
cases significantly fewer lines are visible. If the number of lines
considered is too high, in images showing few straight lines several
false detections are obtained (see left image in Fig.~\ref{fig:Related}).
On the other hand, if the number of lines considered is too low, in
images with many lines (e.g., images showing any of the goal areas),
some lines are misdetected (see right image in Fig.~\ref{fig:Related}).
To reduce these misdetections, some works have proposed the application
of the Hough transform independently along small windows that cover
the entire image~\cite{ali2012efficient}. A further drawback of
Hough-based methods is that the images generally suffer from radial
distortion. Consequently, duplicate lines are typically obtained (see
the bottom line shown in the left image in Fig.~\ref{fig:Related}). 

Once the straight lines have been detected, they are typically classified
according to their tilt~\cite{cuevas2020automatic,sadlier2005event}.
However these analyses are limited to the cases in which the position
of the camera used to acquired the images is known~\cite{jabri2011camera}.
Alternatively, the lines are classified in only two sets (longitudinal
and transverse) that are used to determine two vanishing points~\cite{hayet2005fast,homayounfar2017sports}. 

There are also strategies that focus on detecting the center circle
of the playing field~\cite{aleman2014camera}. Some of them use a
6-dimensional Hough transform to detect ellipses~\cite{mukhopadhyay2015survey}
since, due to the perspective of the images, the center circle is
seen as an ellipse in the images. These strategies have very high
computational and memory requirements~\cite{xu1990new,bergen1991probabilistic}.
Additionally, since the Hough transform is applied on edge or Top-Hat
images, their results are inaccurate because of the presence of players,
billboards, etc. An additional limitation of these algorithms is that
they cannot obtain successful results in images where the center circle
does not appear complete. Alternatively, some authors
have proposed strategies using Least Squares Fitting (LSF) methods~\cite{wu2019efficient}.
However, since they are also very sensitive to the presence of data
that do not belong to the ellipse~\cite{wan2003real}, they require
complex steps to discard data from other lines, players, etc~\cite{cuevas2020automatic}.
Another important limitation of all these strategies is that none
of them is able to detect the circles of the penalty areas.

\section{Watershed-based line mark segmentation\label{sec:Watershed-segmentation}}

\begin{figure}[tbh]
\centering{}\includegraphics[width=.6\columnwidth]{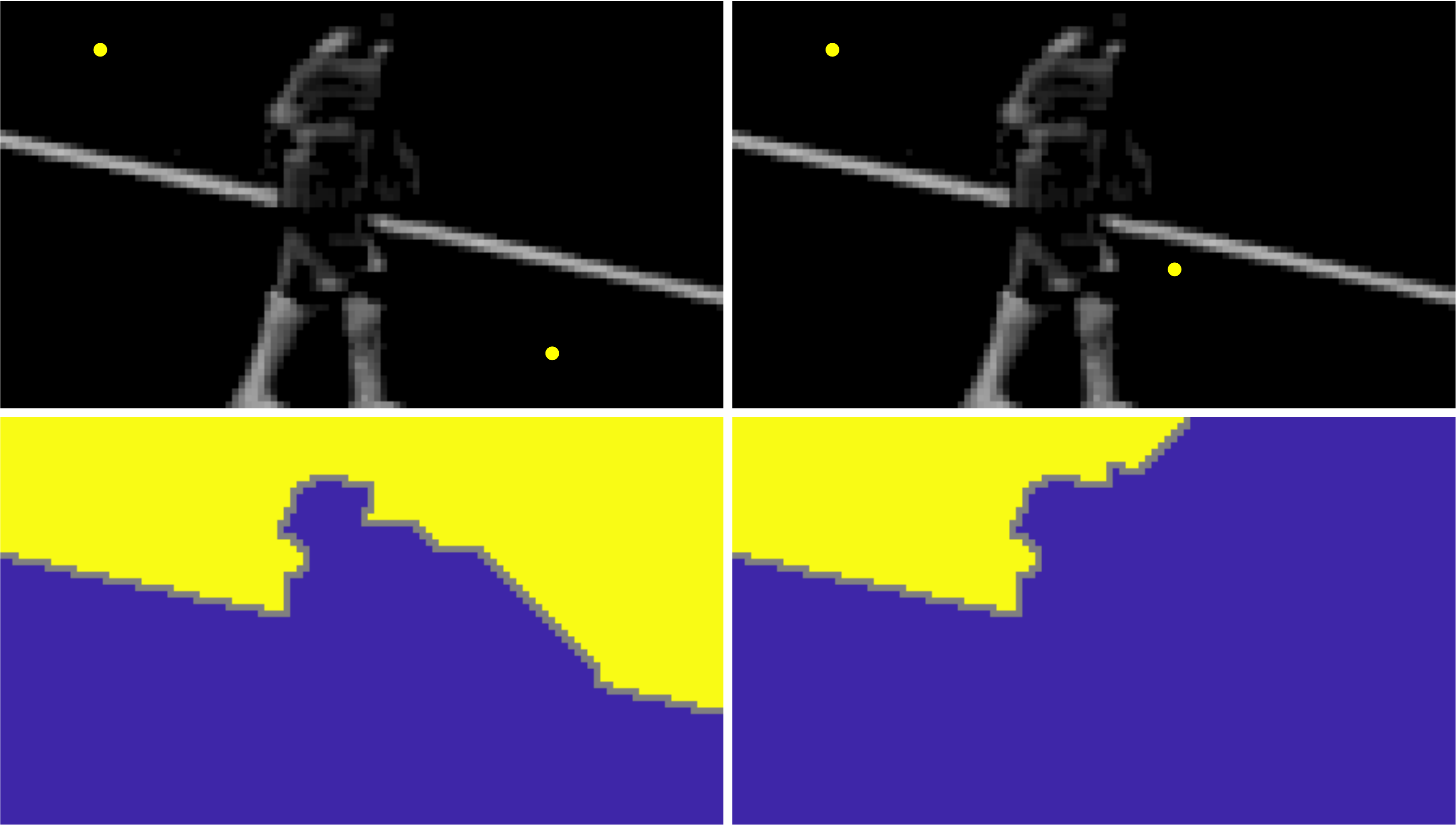}\caption{\label{fig:watershed-marker-position-sensitivity}Examples of different
segmentation results (bottom) depending on the position of the seeds
(top). Note that although a human observer interprets as clearly distinct
areas those above and below the line marking, they are actually joined
due to the interference of the player.}
\end{figure}

We propose a completely automatic procedure based on stochastic watershed~\cite{angulo2007stochasticWatershed,malmberg2014exactStochastic}
to detect the line markings in a soccer pitch. The usual purpose of a watershed
transform is to segment regions in a grayscale image separated by higher ground~\cite{maxwell1870onHillsAndDales}, interpreting the levels the image as 
terrain elevation. However, since every local minimum is such a region, the user has to 
manually determine how many significant regions are there in the image, and even small 
discontinuities in the ridges between regions can upset the results (see Fig.~\ref{fig:watershed-marker-position-sensitivity}).

In our case, however, we are not interested in the regions
themselves but in the lines that delimit them, and we do not care if a
discontinuity in a line as the one shown in Fig.~\ref{fig:watershed-marker-position-sensitivity}
joins two regions.
The basic idea is that, since the line markings are brighter than
their surroundings, if we manage to place \emph{numerous} markers
well distributed throughout both sides of every line mark, the result will
be an image whose watershed lines will include (almost)
every portion of the line markings we want to detect, along with a lot of
spurious lines due to having placed many more markers than there are regions.
However, upon repetition of the experiment with different sets of markers, spurious 
lines will change, as illustrated in Fig.~\ref{fig:watershed-marker-position-sensitivity}, 
but true lines will arise again, and we will reliably detect them by averaging multiple
experiments.

\begin{figure}[tbh]
\centering{}\includegraphics[width=.7\columnwidth]{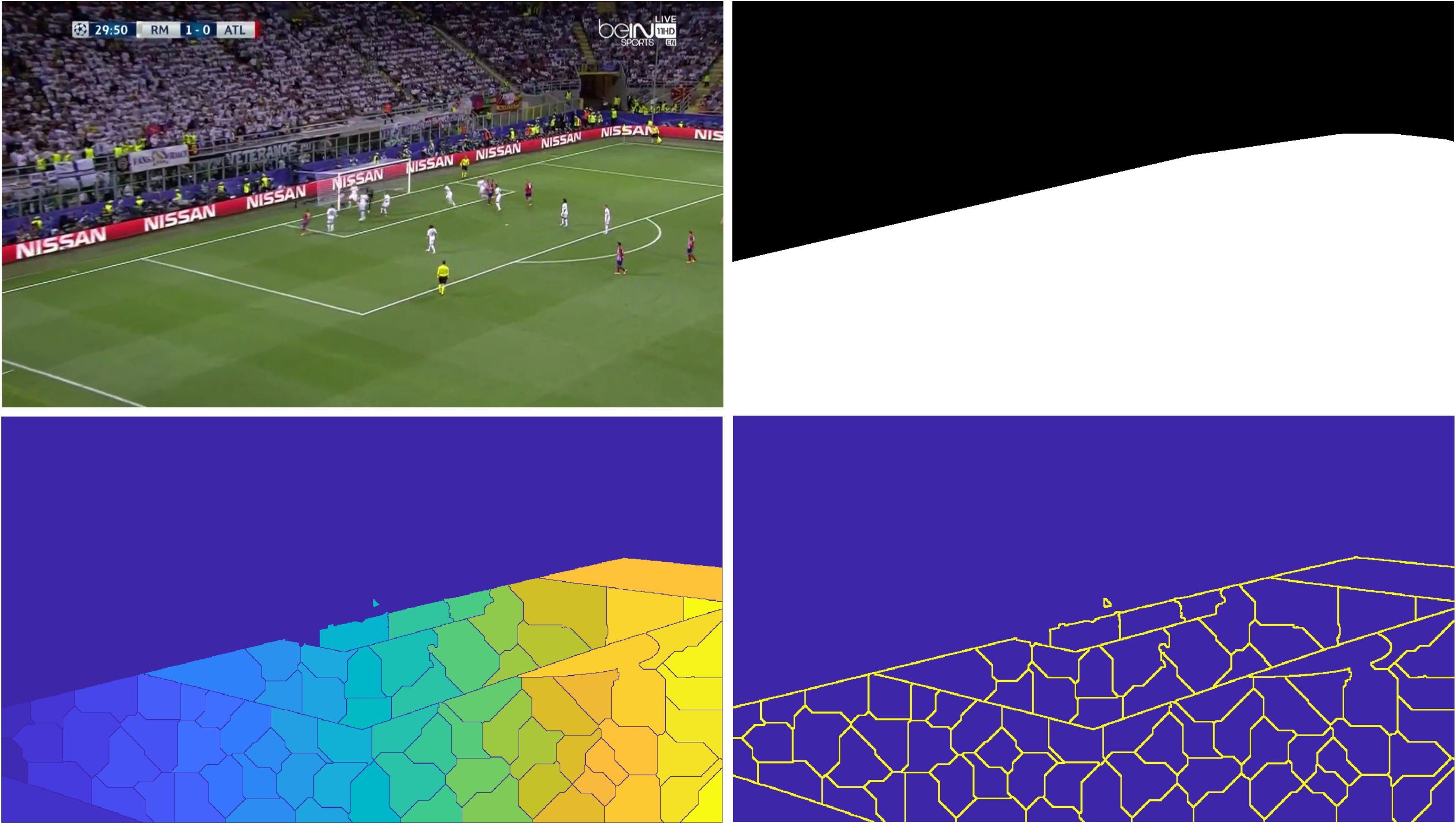}\caption{\label{fig:our-proposal-one-iteration-example}Left to right, top
to bottom: original image, $\protect\imagen$; field of play mask,
$\protect\mascaracampo$; regions determined by a single experiment
of stochastic watershed segmentation; corresponding boundaries.}
\end{figure}

Fig.~\ref{fig:our-proposal-one-iteration-example}
shows an example of a single experiment whose results clearly exhibit
most of the line markings we are looking for, where even lines interrupted
by players (e.g., the penalty arc) are present at both sides of the
interruption.

Let $\maskgrayscale$ be the grayscale image to segment (obtained
as described in section~\ref{sec:Preprocessing}), with a resolution
of $\alto$ rows and $\ancho$ columns. A possible first approach
is to simply generate sets of $\numseeds$ seeds uniformly distributed
across the image:

\begin{equation}
\seedset_{\expindex}=\left\{ \individualseed_{\expindex,\seedindex}=\begin{bmatrix}\rowcoord_{\expindex,\seedindex}\\
\colcoord_{\expindex,\seedindex}
\end{bmatrix}\colon\begin{array}{c}
\rowcoord_{\expindex,\seedindex}=\uniformrv 1{\alto},\\
\colcoord_{\expindex,\seedindex}=\uniformrv 1{\ancho},\\
\seedindex\in\left\{ 1,2,\ldots,\numseeds\right\} 
\end{array}\right\} ,\label{eq:uniform-seed-generation}
\end{equation}
where $\rowcoord$ and $\colcoord$ are the row and column coordinates
where each seed $\individualseed$ is placed, and $\expindex$ identifies
each individual watershed experiment; let us notate the standard seeded
watershed transform on a grayscale image $\maskgrayscale$ with the set of seeds $\seedset$
as $\seededWS{\maskgrayscale}{\seedset}$, yielding a binary image where
the pixels corresponding to watershed lines are set to 1 and all others
are set to 0. Thus, we can compute
\begin{equation}
\randomWatershedImage=\frac{1}{\numexperiments}\sum_{\expindex=1}^{\numexperiments}\seededWS{\maskgrayscale}{\seedset_{\expindex}},
\end{equation}
where $\numexperiments$ is the number of experiments. We can interpret
the value of each pixel of $\randomWatershedImage$ as akin to the
probability of it being a true watershed line of $\maskgrayscale$.
Since we are interested in robust line detection, the final boundary
mask $\mascaralineas$ will only consider as positive detection pixels
with values exceeding a threshold $\thresholdRWS$\footnote{We have used $\thresholdRWS=0.8$ throughout all the reported experiments
to guarantee a robust detection, but it is not a sensitive parameter,
the results are very similar for a wide range of values around the
chosen one.} and located within a region of interest, in our case the playing field, denoted as $\mascaracampo$\footnote{There are many available methods to segment the playing field, i.e.,
the area covered by grass, in the image, namely~\cite{quilon2015unsupervised}.}:

\begin{equation}
\myfunc{\mascaralineas}{\rowcoord,\colcoord}=\begin{cases}
1, & \myfunc{\randomWatershedImage}{\rowcoord,\colcoord}\geq\thresholdRWS\land\myfunc{\mascaracampo}{\rowcoord,\colcoord}=1;\\
0, & \myfunc{\randomWatershedImage}{\rowcoord,\colcoord}<\thresholdRWS\lor\myfunc{\mascaracampo}{\rowcoord,\colcoord}=0.
\end{cases}
\end{equation}

\begin{figure}[tbh]
\begin{centering}
\includegraphics[width=.7\columnwidth]{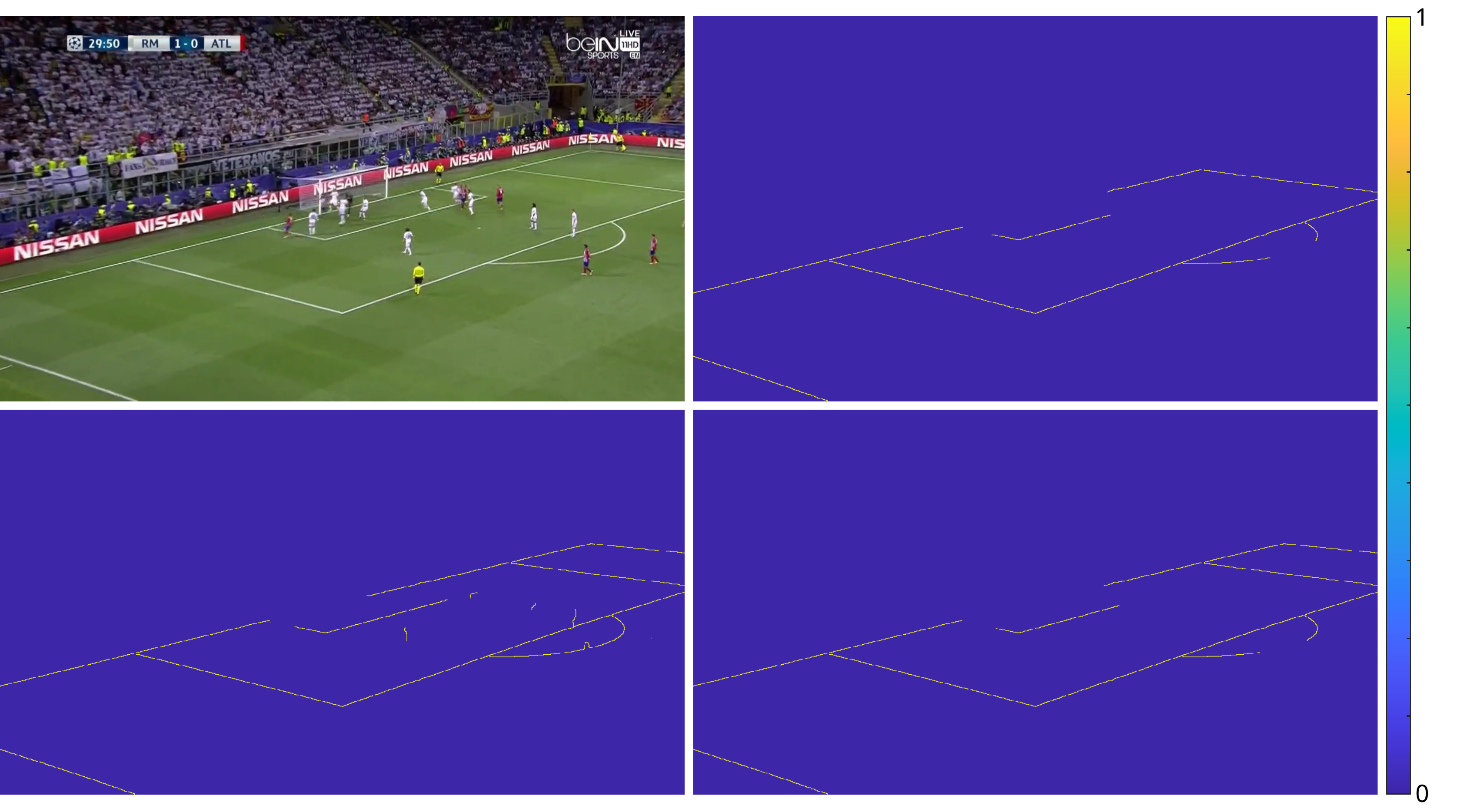}
\par\end{centering}
\caption{\label{fig:uniform-distribution-watershed}Results applying stochastic
watershed with uniform seed distribution for different numbers of
seeds and experiments. Top left: original input; top right: few experiments
and few seeds ($\protect\numexperiments=20,\protect\numseeds=200$);
bottom left: few experiments and many seeds ($\protect\numexperiments=20,\protect\numseeds=1000$);
bottom right: many experiments and few seeds ($\protect\numexperiments=200,\protect\numseeds=200$).}
\end{figure}

However, distributing seeds uniformly across both rows and columns
does not guarantee that every region of the image is covered, as Fig.~\ref{fig:uniform-distribution-watershed}
shows: using a relatively small number of seeds has a non-negligible
chance of leaving regions uncovered, leading to misdetected lines.
Line misdetection can be solved either increasing the number of seeds
or the number of experiments, but the former increases false detections
and, while the latter does produce correct results, it does so at
significantly higher cost.

\begin{figure}[tbh]
\begin{centering}
\includegraphics[width=.7\columnwidth]{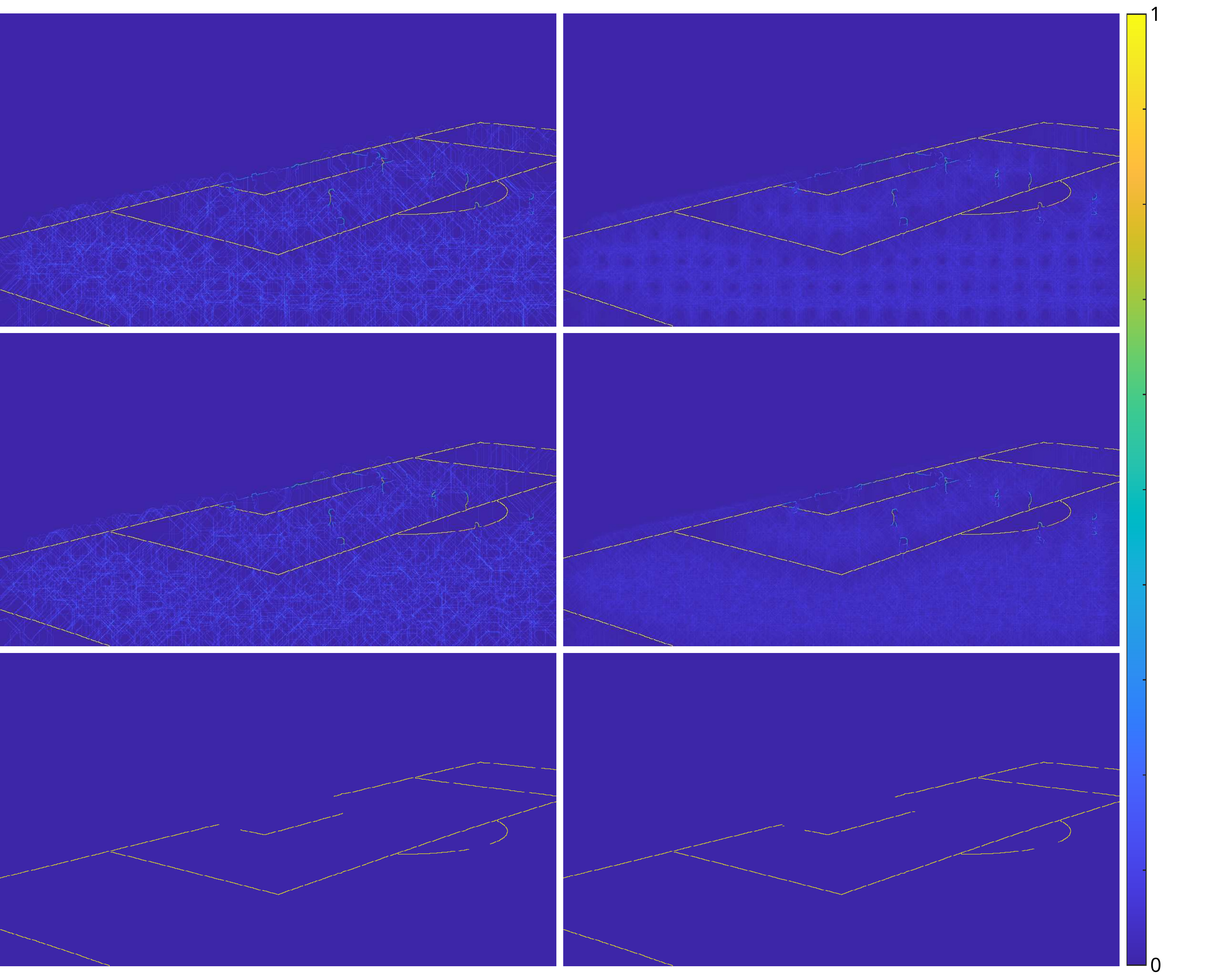}
\par\end{centering}
\caption{\label{fig:Results-old-windows}Results applying stochastic watershed
with windowed random seed generation. On the left, $\protect\numexperiments=20$;
on the right, $\protect\numexperiments=200$. Rows: fixed lattice
across experiments (top), proposed approach with different origin
in each experiment (middle) and proposed approach after thresholding
(bottom).}
\end{figure}

To deal with these problems, we propose a windowed random seed generation
which will ensure that all the regions of the image contain seeds,
attaining quick convergence without increasing the rate of false detections.
Let us divide the input image $\maskgrayscale$ into a lattice of
$\numseedsrows$ vertical divisions and $\numseedscols$ horizontal
divisions that will delimit $\numseedsrows\times\numseedscols$ non-overlapping
rectangular regions (w.l.o.g., let us assume $\alto$ and $\ancho$
are divisible by $\numseedsrows$ and $\numseedscols$ respectively
to simplify notation); then we will place a single seed with uniform
distribution into each of these regions. Thus, we replace equation~\ref{eq:uniform-seed-generation}
with
\begin{equation}
\seedset_{\expindex}=\left\{ \individualseed_{\expindex,\seedindex,k}=\begin{bmatrix}\rowcoord_{\expindex,\seedindex,k}\\
\colcoord_{\expindex,\seedindex,k}
\end{bmatrix}\colon\begin{array}{c}
\rowcoord_{\expindex,\seedindexrow,\seedindexcol}=1+\modulo{\rowcoord_{\mathrm{o},\expindex}+\frac{\seedindexrow\alto}{\numseedsrows}+\uniformrv 1{\frac{\alto}{\numseedsrows}},\alto},\\
\colcoord_{\expindex,\seedindexrow,\seedindexcol}=1+\modulo{\colcoord_{\mathrm{o},\expindex}+\frac{\seedindexcol\ancho}{\numseedscols}+\uniformrv 1{\frac{\ancho}{\numseedscols}},\ancho},\\
\rowcoord_{\mathrm{o},\expindex}=\uniformrv 1{\frac{\alto}{\numseedsrows}},\quad\colcoord_{\mathrm{o},\expindex}=\uniformrv 1{\frac{\ancho}{\numseedscols}},\\
\seedindexrow\in\left\{ 0,1,\ldots,\numseedsrows-1\right\} ,\quad\seedindexcol\in\left\{ 0,1,\ldots,\numseedscols-1\right\} ,
\end{array}\right\} ,
\end{equation}
the rest of the procedure unchanged. Fig.~\ref{fig:Results-old-windows}
shows the results of this seeding method, where we can see that a
small number of experiments ($M=20$) already yields results comparable
to those obtained with the uniform seed generator for $\numexperiments=200$ 
(shown on Fig.~\ref{fig:uniform-distribution-watershed}).
Furthermore, we can also see that increasing significantly the number
of experiments does not yield significant advantages. If the lattice
used to generate seeds has the same origin in every experiment (e.g.,
$\rowcoord_{\mathrm{o},\expindex}=\colcoord_{\mathrm{o},\expindex}=0$),
the procedure yields results which exhibit a faint yet definite pattern,
roughly shaped like the lattice, in flat areas of $\maskgrayscale$,
as illustrated in Fig.~\ref{fig:Results-old-windows}; although this
effect does not actually affect results, it is easy to eliminate by
using a different random origin for the lattice in each experiment,
as proposed.

The proposed seed location algorithm guarantees a bounded distance
between seeds, like the iterative Poisson disk sampling~\cite{bridson2007fastPoisson},
but it has a substantially lower cost and is more amenable to parallelization.

\subsection{Preprocessing\label{sec:Preprocessing}}

As explained before, it should reasonably be possible to use the brightness
of the RGB input image $\imagen$ as the input $\maskgrayscale$ for the 
line detection stage because the line markings have a
higher brightness than their surroundings. However, in practice, this
results in many false line detections due to playing field texture
and capture noise. Therefore, before applying the watershed-based 
line detection we apply a simple pre-processing stage to enhance the 
input image and maximize the contrast of the white lines against the grass.

\begin{figure}[tbh]
\centering{}\includegraphics[width=.7\columnwidth]{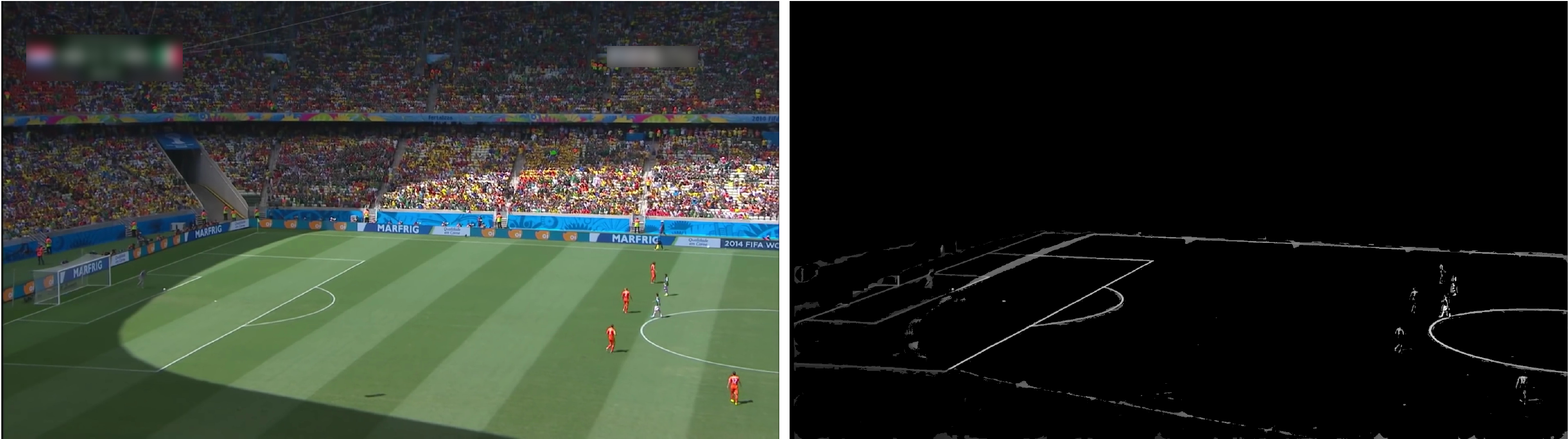}\caption{\label{fig:thresholding-illumination-change-1}Example of false contours
due to abrupt illumination changes when performing a local thresholding
operation (only the contours inside the pitch area are shown).}
\end{figure}

A simple solution is to apply a local thresholding operator~\cite{Bradley2007}
to remove the noisy areas before applying the line detection stage.
However, in the face of abrupt illumination changes this approach
is less than ideal because it may create false contours, which will
then be erroneously detected as lines (see Fig.~\ref{fig:thresholding-illumination-change-1}).
To only preserve thin areas that are lighter than their surroundings in all directions,
we use the Top-Hat transform~\cite{Meyer1979}: $\cosatophat X=X-X\circ\opening$,
where $X$ is a grayscale image, $\circ$ denotes the opening morphological
operation and $\opening$ is a structuring element whose diameter
must be slightly greater than the maximum width of the features to
be preserved (line marks in our case\footnote{In the images of the database we have used to assess the quality of
the strategy (see Section~\ref{sec:Results}) the width of the lines
is typically under 10 pixels. Therefore, a structuring element with
a diameter of 11 pixels has been used.}). The grayscale image to apply the Top-Hat transform onto could just
be the luminance of $\imagen$, but since the line markings are white,
we profit from the fact that they must stand out in each of the R, G and B channels
of $\imagen$. Thus, we use $\maskgrayscale=\myfunc{\min}{\cosatophat R,\cosatophat G,\cosatophat B}$,
so that non-white features are discarded.

\section{Classification of line marks\label{sec:Line-classification}}

\begin{figure*}[tbh]
\begin{centering}
\includegraphics[width=1\textwidth]{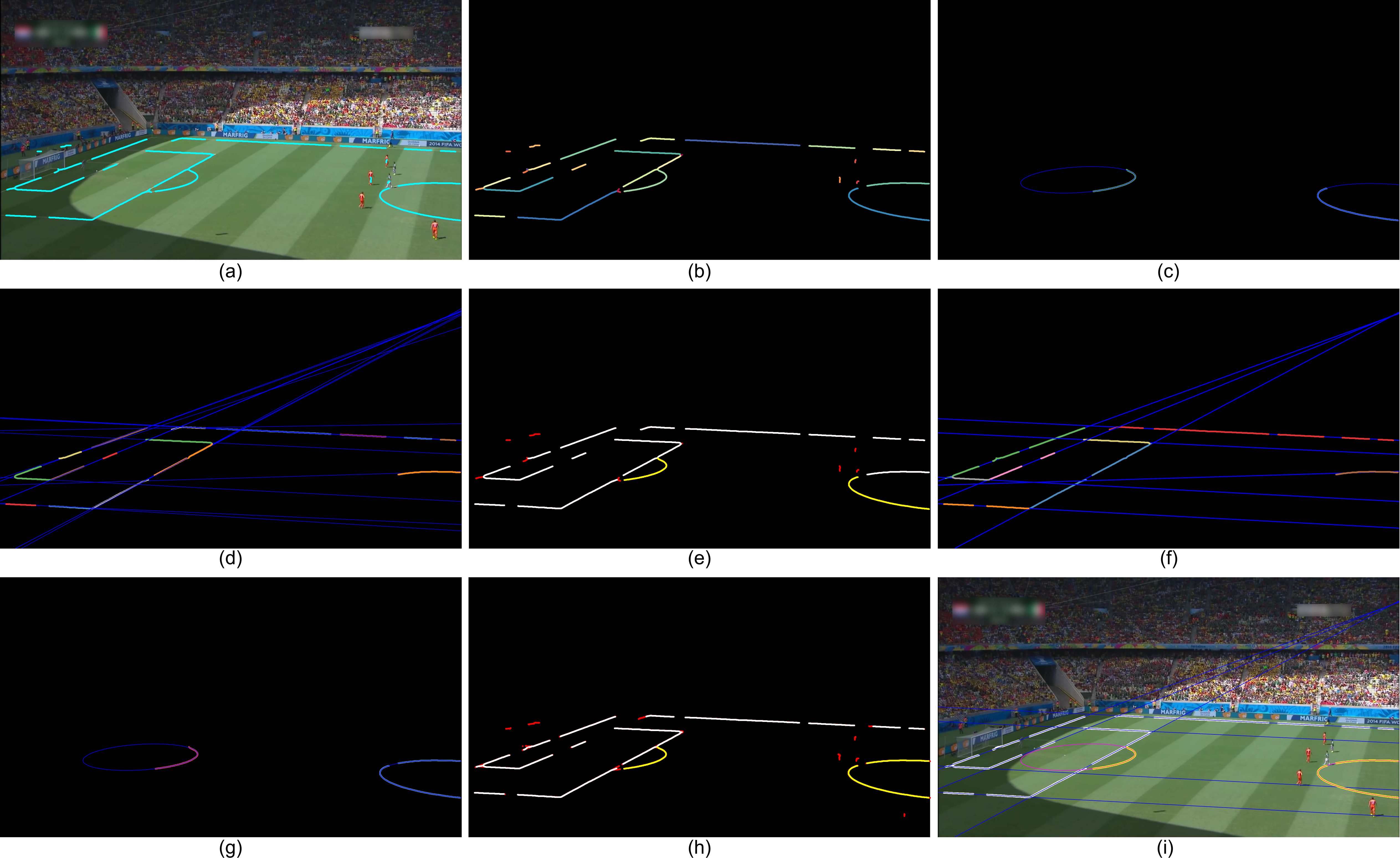}\caption{\label{fig:LineClassification}(a) Original image with detected line
points in cyan. (b) Connected regions (one different color per region).
(c) Ellipses obtained in the initial classification. (d) Straight
lines obtained in the initial classification. (e) Initial classification:
regions associated to ellipses in yellow, regions associated to straight
lines in white, and discarded regions in red. (f) Straight lines and
regions after the straight line merging step (the same color for regions
associated with the same straight line). (g) Ellipses and regions
after the ellipse merging step. (h) Final classification: regions
associated to ellipses in yellow, regions associated to straight lines
in white, and discarded regions in red. (i) Final straight lines with
their associated regions in white and final ellipses with their corresponding
regions in yellow.}
\par\end{centering}
\end{figure*}

The last stage of our strategy is in charge of determining which of
the primitive structures in the playing field (straight line or ellipse)
the detected line points belong to. For this, we have developed a
procedure based on the adjustment of the detected line points to both
types of primitive structures, which consists in the following steps:
\begin{enumerate}
\item Line point segmentation: The line points in $\mascaralineas$ are
segmented into the set of $N_{R}$ connected regions, $R=\{r_{i}\}_{i=1}^{N_{R}}$,
that results after removing line intersections (i.e., line points
with more than two neighbors). As an example, Fig.~\ref{fig:LineClassification}
shows an image with the line points in $\mascaralineas$ (Fig.~\ref{fig:LineClassification}.a)
that has resulted in the set of $N_{R}=26$ connected regions in Fig.~\ref{fig:LineClassification}.b.
\item Initial classification: Each region in $R$ is classified as part
of an ellipse or part of one or more straight lines.
\begin{enumerate}
\item The least squares fitting algorithm in~\cite{fitzgibbon1999direct}
is applied to obtain the ellipse that best fits each connected region.
The root mean square error (RMSE) of this fit is used as a measure
of quality of the fit. 
\item Deming regression~\cite{cornbleet1979incorrect} is used obtain the
straight line that best fits each connected region. If this fit is
not accurate enough, the region is divided into the smallest set of
subregions that allows an accurate fit to a straight line of each
of these subregions. Given that regions corresponding to straight
lines tend to be noisy, we have decided that a fit is accurate enough
when the RMSE of the fit is lower than 2.
\item Each region or subregion is associated with the primitive structure
(ellipse or line) that fits best (i.e., the one with the smallest
RMSE).
\end{enumerate}
Smaller regions (e.g., fewer than 50 pixels) are not representative
enough by themselves. Therefore, they are initially left unassigned
to either primitive structure type.

In the example of Fig.~\ref{fig:LineClassification} it can be seen
that after applying this stage, 2 regions associated with ellipses
(Fig.~\ref{fig:LineClassification}.c) and 18 regions associated
with straight lines (Fig.~\ref{fig:LineClassification}.d) have been
obtained. In addition, in Fig.~\ref{fig:LineClassification}.e it
can be seen that 9 regions have been left unassigned (those depicted
in red). Therefore, in this example we have gone from 26 to 29 regions,
since 3 of the regions have been split to fit two straight lines each.
\item Straight line merging: Regions associated with straight lines are
merged which, after merging, still result in a fine fit. Here, since
there may be lines represented by segments that practically cross
the image, to deal with radial distortion, it has been decided to
relax the criterion by which it is determined if a fit is accurate
enough, allowing a maximum RMSE of 4~px. In this merging step, the small
regions discarded in the previous step are also considered. As it
can be seen in Fig.~\ref{fig:LineClassification}.f, after applying
this step, we have gone from 18 regions represented by 18 lines to
21 regions represented by 8 lines (3 of the previously discarded regions
have been incorporated).
\item Ellipse merging: Regions associated with ellipses are merged if they
still result in a good fit together (the RMSE resulting from the fit
is at most 4). In this step, the possible fusion with regions previously
discarded due to their size is also considered. Notably, not only
unassigned or elliptic regions are considered in this step, but also
regions hitherto thought of as straights are considered because the
curvature of ellipses varies so much that parts of them may locally
fit a straight line (for example, the region corresponding to the
top of the central circle in the example in Fig.~\ref{fig:LineClassification})
and it is only possible to determine their correspondence with elliptical
structures when multiple regions are considered together. Fig.~\ref{fig:LineClassification}.g
shows that after applying this step, the mentioned region has been
associated with the ellipse corresponding to the central circle of
the playing field.
\item Refinement: Ellipses and straight lines are refined by removing pixels
that deviate significantly from the estimated models. The criterion
used to determine which data fit well is the same as those used in
steps 3 and 4 to perform region merging (i.e., the distance to the
model is greater than 4). Fig.~\ref{fig:LineClassification}.h, shows
the final classification of the line points in $\mascaralineas$.
Thanks to this last stage, small segments that do not actually belong
to the geometric model have been discarded (for example, the leftmost
piece of the set of regions associated with the upper touchline).
\end{enumerate}
The result in Fig.~\ref{fig:LineClassification}.i shows that the
proposed strategy has correctly classified most of the line points
detected in the previous stage into their corresponding ellipses (central
circle and arc of the left area) and straight lines. This image also
shows that 7 of the 8 straight lines have been correctly modeled;
the only misdetection, due to its small size in this image, occurs
on the upper line of the goal area.

\section{Results\label{sec:Results}}

\begin{figure*}[htb]
\centering{}\includegraphics[width=1\textwidth]{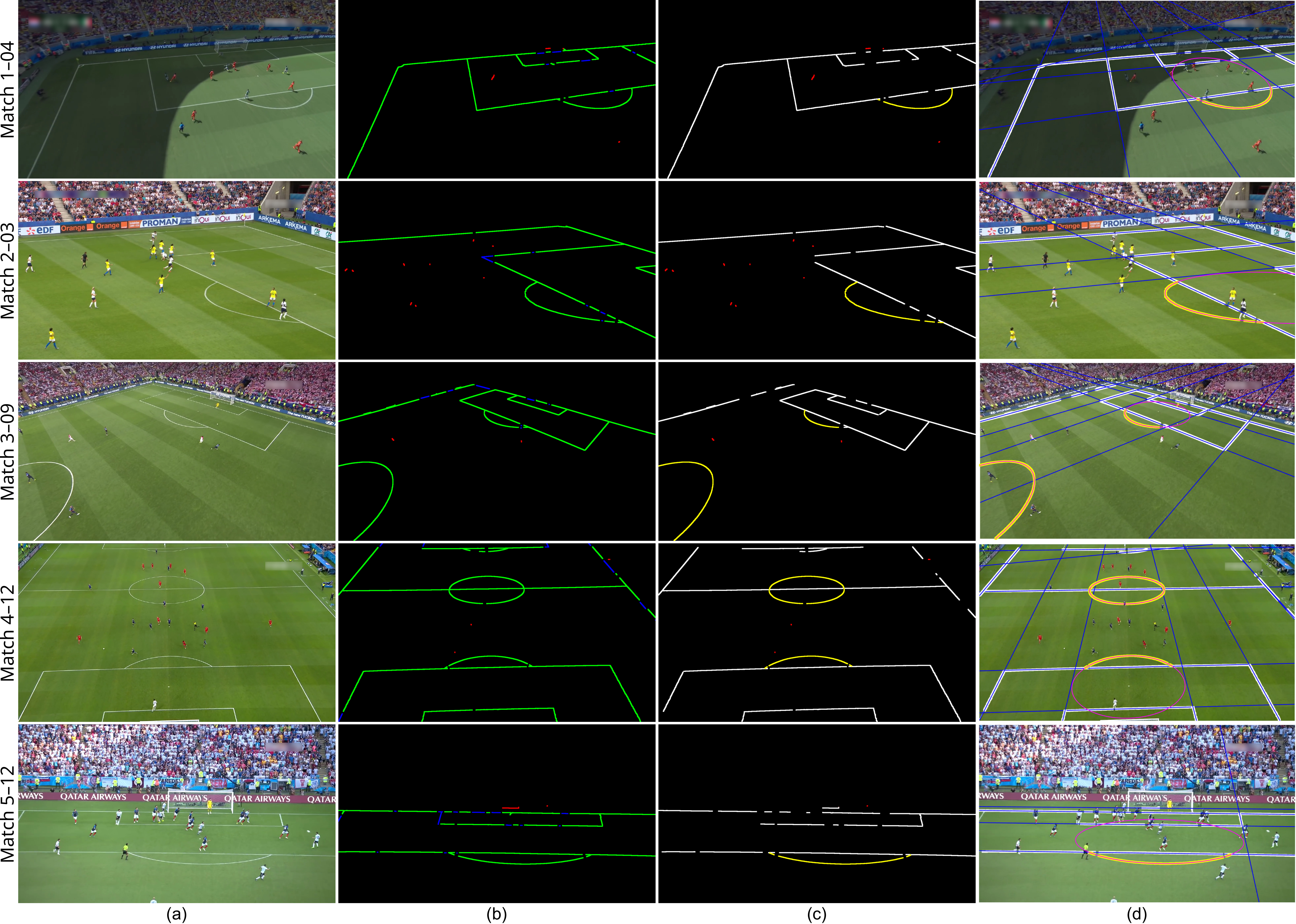}\caption{\label{fig:Some-representative-results}Some representative results
obtained in the test sequences. (a)~Original images. (b)~Line mark
segmentation: true positives in green, false negatives in blue, and
false positives in red. (c)~Line mark classification: in white the
points assigned to straight lines, in yellow the points assigned to
elliptical lines, and in red the discarded points. (d)~Primitives:
straight lines in blue, regions associated to straight lines in white,
ellipses in purple, regions associated to ellipses in yellow.}
\end{figure*}

To assess the quality of the proposed strategy, we have created a
new and public database named LaSoDa\footnote{http://www.gti.ssr.upm.es/data/}.
As far as we know, there is only one other database that, like ours,
allows the evaluation of the quality of algorithms to detect or segment
line marks in images of soccer matches~\cite{homayounfar2017sports}.
However, this database presents two serious shortcomings: all its images 
have been acquired from the same viewpoint and similar wide-view angle, limiting
the variety of its contents; and, crucially, 
many of the homography matrices provided in this database are too 
inaccurate to provide a good registration of a model of the pitch with the images,
precluding a pixel-level evaluation of line mark detections.

LaSoDa is composed by 60 annotated images from matches in five stadiums
with different characteristics (e.g., positions of the cameras, view
angles, grass colors) and light conditions (day and night). Its images
cover all areas of the playing field, show five different zoom levels---from
1 (closest zoom) to 5 (widest zoom)---, have been acquired with four
different types of cameras---master camera (MC), side camera (SC),
end camera (EC), and aerial camera (AC)---, and include different
and challenging lighting conditions (e.g., day and night matches,
and some heavily shaded images).

The quality of the results has been measured by the recall ($\recall$),
precision ($\precision$), and F-score ($\fscore$) metrics, which
are computed as: 

\begin{equation}
\recall=\frac{\truepos}{\truepos+\falseneg},\;\precision=\frac{\truepos}{\truepos+\falsepos},\;\fscore=\frac{2\truepos}{2\truepos+\falsepos+\falseneg},
\end{equation}
where $\truepos$, $\falseneg$, and $\falsepos$ are, respectively,
the amounts of true positives, false negatives and false positives.

{
\setlength{\tabcolsep}{1.5pt}
\renewcommand*{\arraystretch}{1.1}
\footnotesize
\begin{longtable}[c]{|ccc|ccc||ccc|ccc||ccc|ccc|}
\caption{\label{tab:Summary-of-results}Results at pixel and object level} \tabularnewline
\hline 
\multirow{3}{*}{\makecell{Image\\Id.}} & \multirow{3}{*}{\makecell{Camera\\Type}} & \multirow{3}{*}{\makecell{Zoom\\level}} & \multicolumn{3}{c||}{Segmentation} & \multicolumn{6}{c||}{Classification (px)} & \multicolumn{6}{c|}{Classification (obj)}\tabularnewline
\cline{7-18} \cline{8-18} \cline{9-18} \cline{10-18} \cline{11-18} \cline{12-18} \cline{13-18} \cline{14-18} \cline{15-18} \cline{16-18} \cline{17-18} \cline{18-18} 
 &  &  & \multicolumn{3}{c||}{} & \multicolumn{3}{c|}{Straight} & \multicolumn{3}{c||}{Elliptical} & \multicolumn{3}{c|}{Straight} & \multicolumn{3}{c|}{Elliptical}\tabularnewline
\cline{4-18} \cline{5-18} \cline{6-18} \cline{7-18} \cline{8-18} \cline{9-18} \cline{10-18} \cline{11-18} \cline{12-18} \cline{13-18} \cline{14-18} \cline{15-18} \cline{16-18} \cline{17-18} \cline{18-18} 
 &  &  & rec & pre & F & rec & pre & F & rec & pre & F & rec & pre & F & rec & pre & F\tabularnewline
\endfirsthead
\hline 
\multirow{3}{*}{\makecell{Image\\Id.}} & \multirow{3}{*}{\makecell{Camera\\Type}} & \multirow{3}{*}{\makecell{Zoom\\level}} & \multicolumn{3}{c||}{Segmentation} & \multicolumn{6}{c||}{Classification (px)} & \multicolumn{6}{c|}{Classification (obj)}\tabularnewline
\cline{7-18} \cline{8-18} \cline{9-18} \cline{10-18} \cline{11-18} \cline{12-18} \cline{13-18} \cline{14-18} \cline{15-18} \cline{16-18} \cline{17-18} \cline{18-18} 
 &  &  & \multicolumn{3}{c||}{} & \multicolumn{3}{c|}{Straight} & \multicolumn{3}{c||}{Elliptical} & \multicolumn{3}{c|}{Straight} & \multicolumn{3}{c|}{Elliptical}\tabularnewline
\cline{4-18} \cline{5-18} \cline{6-18} \cline{7-18} \cline{8-18} \cline{9-18} \cline{10-18} \cline{11-18} \cline{12-18} \cline{13-18} \cline{14-18} \cline{15-18} \cline{16-18} \cline{17-18} \cline{18-18} 
 &  &  & rec & pre & F & rec & pre & F & rec & pre & F & rec & pre & F & rec & pre & F\tabularnewline
\endhead
\hline 
1--01 & MC & 2 & 0.90 & 0.99 & 0.94 & 0.78 & 1.00 & 0.88 & 1.00 & 1.00 & 1.00 & 0.75 & 1.00 & 0.86 & 1.00 & 1.00 & 1.00\tabularnewline
\hline 
1--02 & MC & 4 & 0.94 & 0.98 & 0.96 & 0.89 & 1.00 & 0.94 & 0.96 & 1.00 & 0.98 & 1.00 & 1.00 & 1.00 & 1.00 & 1.00 & 1.00\tabularnewline
\hline 
1--03 & EC & 1 & 0.93 & 1.00 & 0.96 & 0.81 & 1.00 & 0.90 & 0.85 & 1.00 & 0.92 & 0.75 & 0.82 & 0.78 & 0.67 & 1.00 & 0.80\tabularnewline
\hline 
1--04 & AC & 3 & 0.97 & 1.00 & 0.99 & 0.95 & 1.00 & 0.98 & 1.00 & 1.00 & 1.00 & 1.00 & 1.00 & 1.00 & 1.00 & 1.00 & 1.00\tabularnewline
\hline 
1--05 & MC & 4 & 0.99 & 0.97 & 0.98 & 0.98 & 1.00 & 0.99 & - & - & - & 1.00 & 1.00 & 1.00 & - & - & -\tabularnewline
\hline 
1--06 & MC & 2 & 0.87 & 0.99 & 0.93 & 0.78 & 1.00 & 0.87 & 0.98 & 1.00 & 0.99 & 1.00 & 0.83 & 0.91 & 1.00 & 1.00 & 1.00\tabularnewline
\pagebreak[0]
\hline 
1--07 & MC & 5 & 0.89 & 1.00 & 0.94 & 0.84 & 0.94 & 0.89 & 0.00 & 1.00 & 0.00 & 1.00 & 0.78 & 0.88 & 0.00 & 1.00 & 0.00\tabularnewline
\hline 
1--08 & MC & 3 & 0.96 & 0.99 & 0.97 & 0.95 & 1.00 & 0.97 & - & - & - & 1.00 & 1.00 & 1.00 & - & - & -\tabularnewline
\hline 
1--09 & MC & 3 & 0.97 & 1.00 & 0.98 & 0.96 & 1.00 & 0.98 & 0.76 & 1.00 & 0.86 & 1.00 & 1.00 & 1.00 & 1.00 & 1.00 & 1.00\tabularnewline
\hline 
1--10 & SC & 3 & 0.92 & 0.99 & 0.95 & 0.89 & 1.00 & 0.94 & 1.00 & 1.00 & 1.00 & 1.00 & 1.00 & 1.00 & 1.00 & 1.00 & 1.00\tabularnewline
\hline 
1--11 & AC & 3 & 1.00 & 1.00 & 1.00 & 0.98 & 1.00 & 0.99 & 0.99 & 1.00 & 1.00 & 1.00 & 1.00 & 1.00 & 1.00 & 1.00 & 1.00\tabularnewline
\hline 
1--12 & MC & 4 & 0.94 & 1.00 & 0.97 & 0.89 & 1.00 & 0.94 & 0.95 & 1.00 & 0.98 & 1.00 & 0.86 & 0.92 & 1.00 & 1.00 & 1.00\tabularnewline
\pagebreak[1]
\hline 
2--01 & MC & 3 & 0.97 & 0.98 & 0.98 & 0.91 & 0.96 & 0.93 & 1.00 & 1.00 & 1.00 & 1.00 & 1.00 & 1.00 & 1.00 & 1.00 & 1.00\tabularnewline
\hline 
2--02 & MC & 3 & 0.96 & 0.99 & 0.98 & 0.94 & 0.96 & 0.95 & 0.00 & 1.00 & 0.00 & 1.00 & 0.80 & 0.89 & 0.00 & 1.00 & 0.00\tabularnewline
\hline 
2--03 & MC & 3 & 0.97 & 0.99 & 0.98 & 0.95 & 1.00 & 0.97 & 0.99 & 1.00 & 1.00 & 1.00 & 0.86 & 0.92 & 1.00 & 1.00 & 1.00\tabularnewline
\hline 
2--04 & EC & 4 & 0.98 & 0.98 & 0.98 & 0.97 & 1.00 & 0.98 & 0.97 & 1.00 & 0.98 & 1.00 & 1.00 & 1.00 & 1.00 & 1.00 & 1.00\tabularnewline
\hline 
2--05 & SC & 3 & 0.99 & 0.99 & 0.99 & 0.98 & 0.99 & 0.99 & 0.96 & 1.00 & 0.98 & 1.00 & 0.90 & 0.95 & 1.00 & 1.00 & 1.00\tabularnewline
\hline 
2--06 & SC & 3 & 0.98 & 0.99 & 0.99 & 0.97 & 1.00 & 0.98 & 0.98 & 1.00 & 0.99 & 1.00 & 1.00 & 1.00 & 1.00 & 1.00 & 1.00\tabularnewline
\pagebreak[0]
\hline 
2--07 & MC & 3 & 0.93 & 1.00 & 0.96 & 0.90 & 1.00 & 0.95 & 0.94 & 1.00 & 0.97 & 1.00 & 0.67 & 0.80 & 1.00 & 1.00 & 1.00\tabularnewline
\hline 
2--08 & MC & 2 & 0.95 & 0.99 & 0.97 & 0.96 & 0.96 & 0.96 & 0.00 & 1.00 & 0.00 & 1.00 & 0.80 & 0.89 & 0.00 & 1.00 & 0.00\tabularnewline
\hline 
2--09 & SC & 5 & 0.82 & 0.97 & 0.89 & 0.79 & 1.00 & 0.88 & - & - & - & 1.00 & 0.83 & 0.91 & - & - & -\tabularnewline
\hline 
2--10 & SC & 5 & 0.76 & 0.97 & 0.85 & 0.70 & 1.00 & 0.83 & - & - & - & 1.00 & 0.71 & 0.83 & - & - & -\tabularnewline
\hline 
2--11 & EC & 4 & 0.97 & 0.99 & 0.98 & 0.94 & 0.88 & 0.91 & 0.58 & 1.00 & 0.73 & 0.67 & 0.44 & 0.53 & 0.50 & 1.00 & 0.67\tabularnewline
\hline 
2--12 & SC & 1 & 0.96 & 0.97 & 0.97 & 0.93 & 0.97 & 0.95 & 0.94 & 1.00 & 0.97 & 1.00 & 1.00 & 1.00 & 1.00 & 1.00 & 1.00\tabularnewline
\pagebreak[1]
\hline 
3--01 & EC & 1 & 0.93 & 0.98 & 0.95 & 0.91 & 0.99 & 0.95 & 0.96 & 1.00 & 0.98 & 1.00 & 0.75 & 0.86 & 1.00 & 1.00 & 1.00\tabularnewline
\hline 
3--02 & MC & 2 & 0.99 & 0.99 & 0.99 & 0.97 & 1.00 & 0.99 & 1.00 & 1.00 & 1.00 & 1.00 & 0.90 & 0.95 & 1.00 & 1.00 & 1.00\tabularnewline
\hline 
3--03 & MC & 4 & 0.97 & 1.00 & 0.98 & 0.93 & 1.00 & 0.97 & - & - & - & 1.00 & 1.00 & 1.00 & - & - & -\tabularnewline
\hline 
3--04 & MC & 5 & 0.94 & 1.00 & 0.97 & 0.91 & 0.92 & 0.92 & 0.00 & 1.00 & 0.00 & 1.00 & 0.89 & 0.94 & 0.00 & 1.00 & 0.00\tabularnewline
\hline 
3--05 & MC & 1 & 0.99 & 0.99 & 0.99 & 0.99 & 1.00 & 1.00 & 0.99 & 1.00 & 0.99 & 1.00 & 1.00 & 1.00 & 1.00 & 1.00 & 1.00\tabularnewline
\hline 
3--06 & MC & 4 & 0.95 & 0.98 & 0.96 & 0.94 & 1.00 & 0.97 & 0.94 & 1.00 & 0.97 & 1.00 & 1.00 & 1.00 & 1.00 & 1.00 & 1.00\tabularnewline
\pagebreak[0]
\hline 
3--07 & AC & 3 & 1.00 & 1.00 & 1.00 & 0.96 & 1.00 & 0.98 & 1.00 & 1.00 & 1.00 & 1.00 & 0.73 & 0.84 & 1.00 & 1.00 & 1.00\tabularnewline
\hline 
3--08 & MC & 1 & 0.99 & 0.99 & 0.99 & 0.99 & 1.00 & 1.00 & 0.96 & 1.00 & 0.98 & 1.00 & 1.00 & 1.00 & 1.00 & 1.00 & 1.00\tabularnewline
\hline 
3--09 & AC & 1 & 1.00 & 1.00 & 1.00 & 1.00 & 1.00 & 1.00 & 0.98 & 1.00 & 0.99 & 1.00 & 0.89 & 0.94 & 1.00 & 1.00 & 1.00\tabularnewline
\hline 
3--10 & SC & 5 & 0.97 & 0.99 & 0.98 & 0.95 & 0.96 & 0.96 & 0.00 & 1.00 & 0.00 & 1.00 & 0.82 & 0.90 & 0.00 & 1.00 & 0.00\tabularnewline
\hline 
3--11 & AC & 4 & 0.98 & 1.00 & 0.99 & 0.97 & 1.00 & 0.98 & 0.95 & 1.00 & 0.97 & 1.00 & 1.00 & 1.00 & 1.00 & 1.00 & 1.00\tabularnewline
\hline 
3--12 & AC & 1 & 0.98 & 1.00 & 0.99 & 0.97 & 1.00 & 0.98 & 0.99 & 1.00 & 0.99 & 0.78 & 1.00 & 0.88 & 1.00 & 1.00 & 1.00\tabularnewline
\pagebreak[1]
\hline 
4--01 & AC & 1 & 0.99 & 1.00 & 1.00 & 0.99 & 0.95 & 0.97 & 0.74 & 1.00 & 0.85 & 0.78 & 0.78 & 0.78 & 0.50 & 1.00 & 0.67\tabularnewline
\hline 
4--02 & AC & 3 & 1.00 & 1.00 & 1.00 & 0.99 & 1.00 & 1.00 & 1.00 & 1.00 & 1.00 & 1.00 & 1.00 & 1.00 & 1.00 & 1.00 & 1.00\tabularnewline
\hline 
4--03 & AC & 3 & 0.99 & 1.00 & 0.99 & 0.99 & 1.00 & 0.99 & 1.00 & 1.00 & 1.00 & 1.00 & 1.00 & 1.00 & 1.00 & 1.00 & 1.00\tabularnewline
\hline 
4--04 & SC & 3 & 0.94 & 0.99 & 0.96 & 0.93 & 1.00 & 0.96 & 0.90 & 1.00 & 0.95 & 1.00 & 0.89 & 0.94 & 1.00 & 1.00 & 1.00\tabularnewline
\hline 
4--05 & AC & 3 & 0.98 & 0.99 & 0.99 & 0.97 & 1.00 & 0.98 & 0.99 & 1.00 & 1.00 & 1.00 & 1.00 & 1.00 & 1.00 & 1.00 & 1.00\tabularnewline
\hline 
4--06 & SC & 2 & 0.98 & 0.99 & 0.99 & 0.97 & 1.00 & 0.98 & 0.97 & 1.00 & 0.99 & 0.80 & 1.00 & 0.89 & 1.00 & 1.00 & 1.00\tabularnewline
\pagebreak[0]
\hline 
4--07 & MC & 3 & 0.99 & 1.00 & 0.99 & 0.94 & 1.00 & 0.97 & 0.97 & 1.00 & 0.99 & 1.00 & 1.00 & 1.00 & 1.00 & 1.00 & 1.00\tabularnewline
\hline 
4--08 & AC & 4 & 0.94 & 1.00 & 0.97 & 0.96 & 0.98 & 0.97 & 0.00 & 1.00 & 0.00 & 1.00 & 0.67 & 0.80 & 0.00 & 1.00 & 0.00\tabularnewline
\hline 
4--09 & AC & 1 & 0.95 & 0.96 & 0.95 & 0.92 & 0.98 & 0.95 & 0.99 & 1.00 & 1.00 & 1.00 & 0.73 & 0.84 & 1.00 & 1.00 & 1.00\tabularnewline
\hline 
4--10 & AC & 3 & 1.00 & 1.00 & 1.00 & 0.98 & 1.00 & 0.99 & 1.00 & 1.00 & 1.00 & 0.86 & 1.00 & 0.92 & 1.00 & 1.00 & 1.00\tabularnewline
\hline 
4--11 & MC & 2 & 0.98 & 1.00 & 0.99 & 0.98 & 1.00 & 0.99 & 0.96 & 1.00 & 0.98 & 1.00 & 0.89 & 0.94 & 1.00 & 1.00 & 1.00\tabularnewline
\hline 
4--12 & AC & 1 & 0.99 & 0.99 & 0.99 & 0.97 & 0.98 & 0.98 & 0.85 & 1.00 & 0.92 & 1.00 & 0.91 & 0.95 & 0.67 & 1.00 & 0.80\tabularnewline
\pagebreak[1]
\hline 
5--01 & MC & 2 & 0.99 & 0.99 & 0.99 & 0.99 & 1.00 & 0.99 & 0.95 & 1.00 & 0.97 & 1.00 & 1.00 & 1.00 & 1.00 & 1.00 & 1.00\tabularnewline
\hline 
5--02 & MC & 3 & 0.95 & 1.00 & 0.98 & 0.93 & 1.00 & 0.96 & 0.96 & 1.00 & 0.98 & 1.00 & 1.00 & 1.00 & 1.00 & 1.00 & 1.00\tabularnewline
\hline 
5--03 & MC & 3 & 0.98 & 0.98 & 0.98 & 0.97 & 1.00 & 0.98 & 0.96 & 0.99 & 0.97 & 1.00 & 1.00 & 1.00 & 1.00 & 1.00 & 1.00\tabularnewline
\hline 
5--04 & AC & 3 & 0.99 & 1.00 & 0.99 & 0.98 & 1.00 & 0.99 & 0.99 & 1.00 & 0.99 & 1.00 & 1.00 & 1.00 & 1.00 & 1.00 & 1.00\tabularnewline
\hline 
5--05 & AC & 5 & 0.99 & 1.00 & 1.00 & 0.99 & 1.00 & 1.00 & - & - & - & 1.00 & 1.00 & 1.00 & - & - & -\tabularnewline
\hline 
5--06 & MC & 5 & 0.89 & 1.00 & 0.94 & 0.86 & 1.00 & 0.92 & - & - & - & 0.80 & 0.80 & 0.80 & - & - & -\tabularnewline
\pagebreak[0]
\hline 
5--07 & SC & 5 & 0.78 & 0.99 & 0.87 & 0.75 & 0.99 & 0.85 & 0.00 & 1.00 & 0.00 & 0.71 & 0.83 & 0.77 & 0.00 & 1.00 & 0.00\tabularnewline
\hline 
5--08 & AC & 5 & 0.80 & 0.99 & 0.89 & 0.74 & 1.00 & 0.85 & 0.82 & 0.97 & 0.89 & 0.71 & 0.83 & 0.77 & 1.00 & 1.00 & 1.00\tabularnewline
\hline 
5--09 & SC & 4 & 0.90 & 1.00 & 0.94 & 0.90 & 1.00 & 0.95 & 0.54 & 1.00 & 0.70 & 1.00 & 0.89 & 0.94 & 1.00 & 1.00 & 1.00\tabularnewline
\hline 
5--10 & SC & 3 & 0.88 & 1.00 & 0.93 & 0.87 & 1.00 & 0.93 & 0.57 & 1.00 & 0.72 & 1.00 & 1.00 & 1.00 & 1.00 & 1.00 & 1.00\tabularnewline
\hline 
5--11 & AC & 1 & 0.97 & 0.99 & 0.98 & 0.95 & 1.00 & 0.97 & 0.95 & 1.00 & 0.97 & 1.00 & 1.00 & 1.00 & 1.00 & 1.00 & 1.00\tabularnewline
\hline 
5--12 & AC & 4 & 0.93 & 1.00 & 0.97 & 0.91 & 1.00 & 0.95 & 1.00 & 1.00 & 1.00 & 0.67 & 1.00 & 0.80 & 1.00 & 1.00 & 1.00\tabularnewline
\hline 
\multicolumn{3}{|c|}{Match 1} & 0.94 & 0.99 & 0.96 & 0.89 & 0.99 & 0.94 & 0.92 & 1.00 & 0.96 & 0.94 & 0.93 & 0.93 & 0.86 & 1.00 & 0.92\tabularnewline
\hline 
\multicolumn{3}{|c|}{Match 2} & 0.95 & 0.99 & 0.97 & 0.92 & 0.97 & 0.94 & 0.82 & 1.00 & 0.90 & 0.97 & 0.81 & 0.88 & 0.77 & 1.00 & 0.87\tabularnewline
\hline 
\multicolumn{3}{|c|}{Match 3} & 0.97 & 0.99 & 0.98 & 0.96 & 0.99 & 0.97 & 0.89 & 1.00 & 0.94 & 0.98 & 0.89 & 0.93 & 0.88 & 1.00 & 0.93\tabularnewline
\hline 
\multicolumn{3}{|c|}{Match 4} & 0.98 & 0.99 & 0.98 & 0.97 & 0.99 & 0.98 & 0.89 & 1.00 & 0.94 & 0.95 & 0.88 & 0.92 & 0.81 & 1.00 & 0.90\tabularnewline
\hline 
\multicolumn{3}{|c|}{Match 5} & 0.92 & 0.99 & 0.96 & 0.90 & 1.00 & 0.94 & 0.89 & 1.00 & 0.94 & 0.91 & 0.94 & 0.92 & 0.92 & 1.00 & 0.96\tabularnewline
\hline 
\multicolumn{3}{|c|}{Total} & 0.95 & 0.99 & 0.97 & 0.93 & 0.99 & 0.96 & 0.88 & 1.00 & 0.93 & 0.95 & 0.89 & 0.92 & 0.85 & 1.00 & 0.92\tabularnewline
\hline 
\end{longtable}

\normalsize
}

\begin{figure*}[tbh]
\centering{}\includegraphics[width=.9\textwidth]{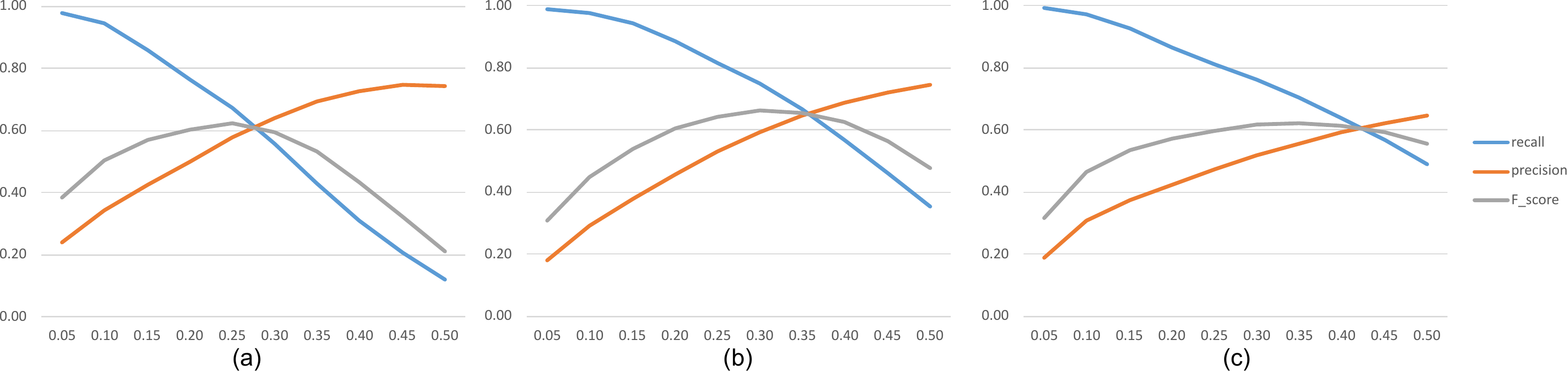}\caption{\label{fig: Segmentation algorithms}Global quality results obtained
with other state-of-the-art line mark segmentation algorithms: (a)
Top-Hat transform, (b) Sobel edge detector, and (c) LoG edge detector.}
\end{figure*}

\begin{figure}[htb]
\centering{}\includegraphics[width=.68\columnwidth]{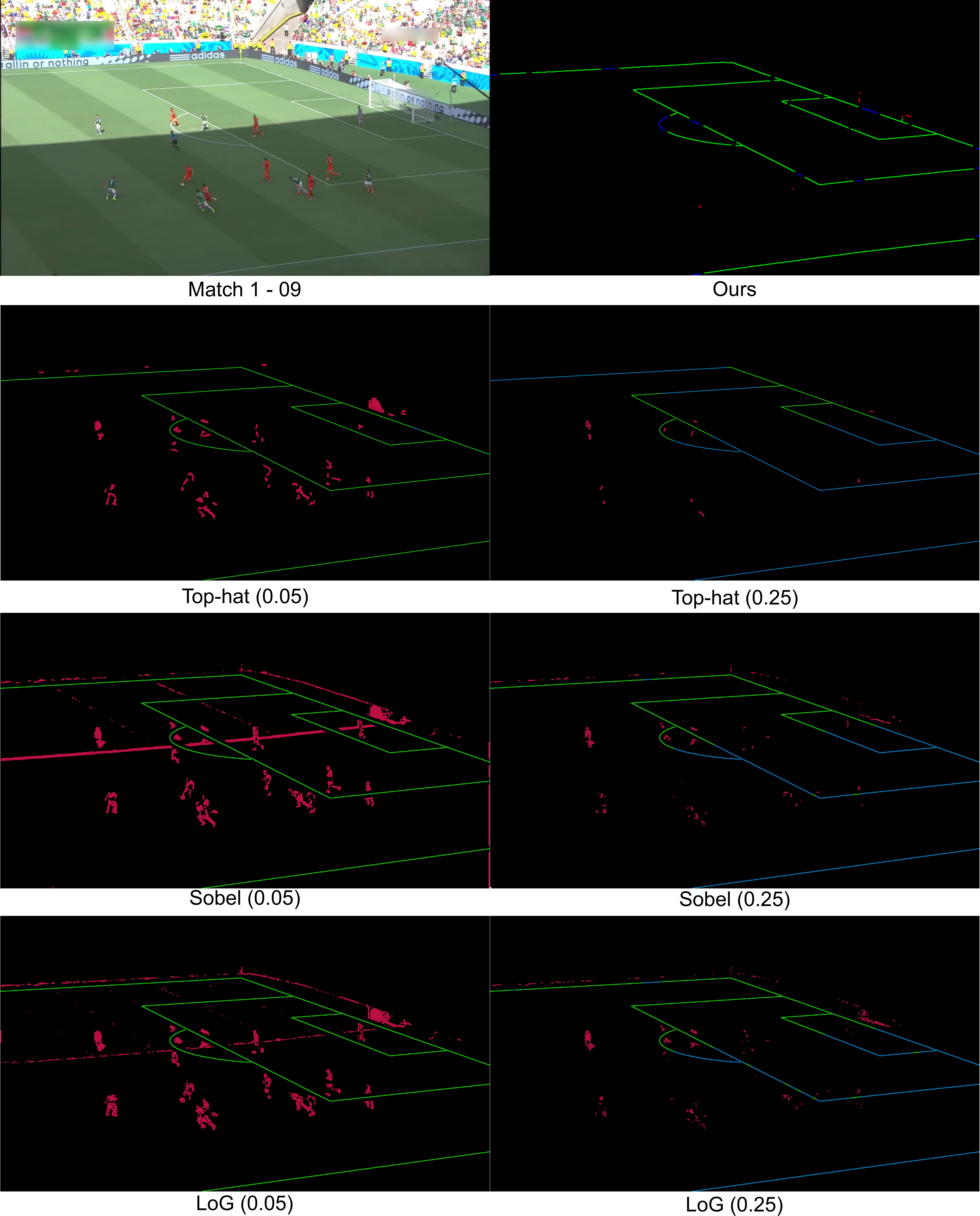}
\caption{\label{fig:Example-of-line}Example of line mark segmentation with
the considered algorithms with different normalized threshold values:
true positives in green, false negatives in blue, and false positives
in red.}
\end{figure}

Table~\ref{tab:Summary-of-results} 
summarizes the quality measures obtained in each of the images in the
proposed database using the provided ground truth for the mask of
the playing field $\mascaracampo$, as well as the
overall quality results for each match individually and for
the entire database. In addition, Fig.~\ref{fig:Some-representative-results}
shows the results of a significant image in each of the five matches
in the database (the results corresponding to the
rest of the images are available on the LaSoDa website).

The ``Segmentation'' columns of Table~\ref{tab:Summary-of-results} 
report the pixel-level results of the
segmentation of the line marks in bulk (i.e., before classification
into straight or elliptical marks). These results show that the quality
of the line mark segmentation is very high, with F-score values over
0.9 in most images and over 0.8 in the remaining few, yielding an
overall F-score of 0.97. The few false positives are mainly due to
players wearing white kits and to white elements of the goals, which
show a similar appearance to that of the white lines on the pitch.
False negatives correspond to line points occluded by players or that
barely stand out from the grass because they are very far from the
camera. 

To compare with the most commonly used methods for line detection, 
Fig.~\ref{fig: Segmentation algorithms} summarizes the global
quality results obtained with three of the most popular algorithms
that are used in other works, as indicated in section~\ref{sec:Related-work},
to highlight the line markings in soccer images: the Sobel edge detector,
the LoG edge detector, and the Top-Hat transform. We can see that none of them
performs well across the whole dataset:
there is no threshold value that gives a really good balance between
precision and recall in any of these methods. Fig.~\ref{fig:Example-of-line}
compares the results obtained on an image where approximately half
of the playing field is strongly sunlit, while the other half is in
shadow. It can be seen that only our strategy is able to
detect the line marks in all areas of the playing field, while at the 
same time yielding the fewest false positives.

The ``Classification (px)'' columns of Table~\ref{tab:Summary-of-results} also 
report pixel-wise results, but after classification into straight 
and elliptical lines, showing that
after applying the line mark classification stage the quality obtained
is very high too. As it can be seen in Fig.~\ref{fig:Some-representative-results} 
and in the table,
most false detections that resulted from the segmentation stage have
been discarded. However, some of the correctly segmented points at
this stage have also been discarded, not being assigned to straight
lines or ellipses, which has caused a slight decrease in the recall.
This generally occurs with line segments that, because of their small
size, have not been conclusively classified as any of the primitive
structures.

Finally, the ``Classification (obj)'' columns of Table~\ref{tab:Summary-of-results}
report the object-level results corresponding to the primitive structures associated with
the line points resulting from the classification
(illustrated in the last column of Fig.~\ref{fig:Some-representative-results}). 
Out of the 402 straight lines featured in the whole of the database,
only 22 have been misdetected. These non-detections are associated
with the shortest line segments, as illustrated in Fig.~\ref{fig:Some-representative-results}.
On the other hand, 59 false detections have been obtained, which are
overwhelmingly due to cases where only the straightest parts of ellipses
(mainly on the penalty arc, which is small and frequently presents
partial occlusion from players) are visible and have been wrongly
classified as small straight lines. Regarding ellipses, out of the
71 that appear in the 60 images analyzed, 12 have been misdetected
but there are no false detections at all. These non-detections are
associated with images where the detected line segments have not been
long enough to determine that they fit an ellipse better than a straight
line (an example of this can also be seen in the last of the images
in Fig.~\ref{fig:Some-representative-results}).

\begin{figure}[tbh]
\centering{}\includegraphics[width=.5\columnwidth]{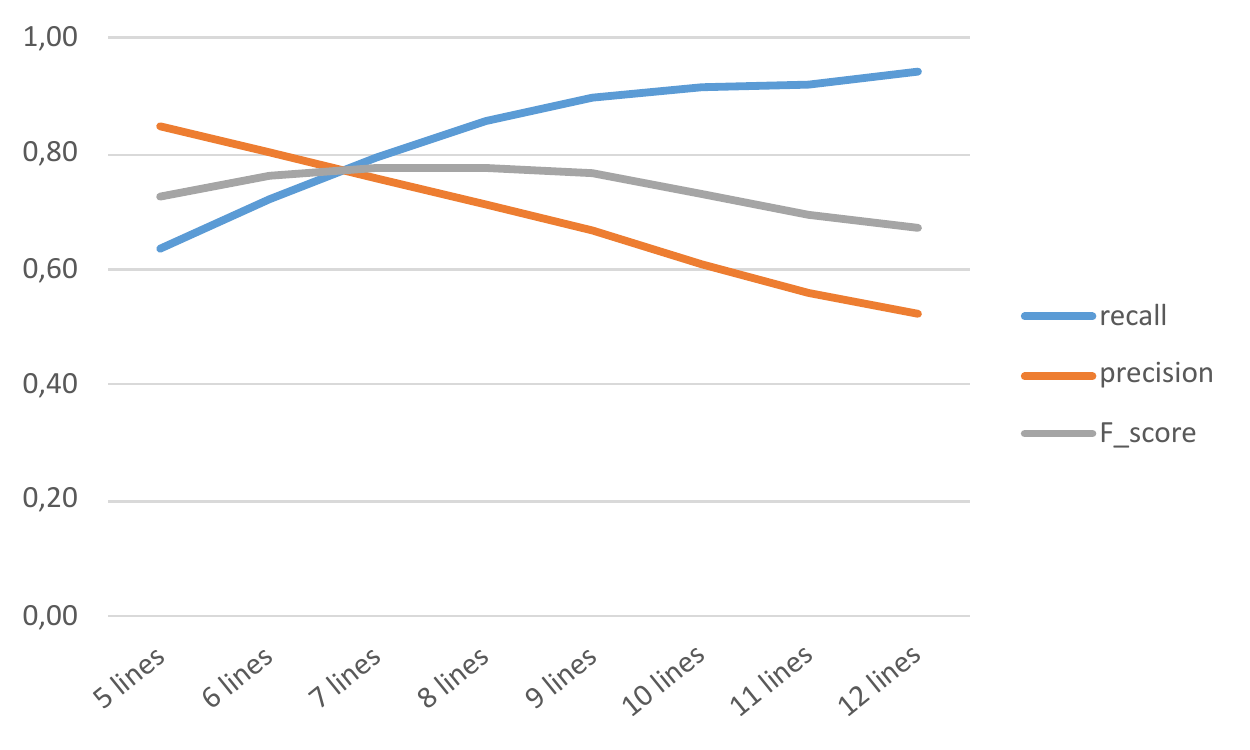}\caption{\label{fig:Quality-results-obtained}Straight line detection results
obtained with the Hough transform.}
\end{figure}

Figure~\ref{fig:Quality-results-obtained} reports the global results
using the Hough transform, a very popular method for straight line
detection in the methods reviewed in section~\ref{sec:Related-work},
from the points in $\mascaralineas$. As already stated in section~\ref{sec:Related-work},
this method requires the user to set a target number of lines to search
for. These results show that setting a low number of target lines
results in many misdetections; conversely, a high number of target
lines results in a significant increase in the number of false detections.
The best result has been obtained by configuring the method to obtain
8 lines per image. However, this global optimal configuration still
results in a relatively low F-score of 0.78, while our proposal reaches
0.90.

\section{Conclusions\label{sec:Conclusions}}

We have presented a novel strategy to segment and
classify line markings in football pitches, irrespectively of camera
position or orientation, and validated its effectiveness on a variety
of images from several stadiums. This strategy is based on the stochastic
watershed transform, which is usually employed to segment regions
rather than lines, but we have shown that, coupled with the seeding
strategy we propose, it provides a robust way to segment the line
markings from the playing field without assuming that the lines are
straight or conform to any particular pattern. Specifically, our method
is able to correctly segment the curved lines that are typical of
wide-angle takes due to radial distortion of broadcasting cameras
or cope with the interference of players or the ball, which frequently
cause errors in most methods based on the Hough transform. Following
the segmentation of the lines, we have also proposed a simple method
to classify the segmented pixels into straight lines and ellipses
or parts thereof. This method is also robust against moderate radial
distortion and outliers, and does not make assumptions about the viewpoint
or the distribution of the markings, leaving those considerations
for later higher-level classification and camera resection stages.

To assess the quality of the proposed strategy, a
new and public database (LaSoDa) has been developed. LaSoDa is composed
of 60 annotated images from matches in stadiums with different characteristics
(positions of the cameras, grass colors) and light conditions (natural
and artificial).

Some improvements can be considered for future work,
for instance parallelization of the watershed transform, since the
regular arrangement of seeds may allow to independently compute different
regions of the image, but the current proposal already constitutes
a solid foundation for other stages in a sports video analysis pipeline
to build upon.

\section*{Acknowledgments}
This work has been partially supported by MCIN/AEI/10.13039/501100011033 of the Spanish Government [grant number PID2020-115132RB (SARAOS)].

\bibliographystyle{elsarticle-num} 
\bibliography{References}

\end{document}